# TASK TREE RETRIEVAL FOR ROBOTIC COOKING

Chakradhar Reddy Nallu
*Department of Computer Science*
*University Of South Florida*
Tampa, USA
nalluchakradhar@gmail.com

**Abstract:**

This paper is based on developing different algorithms, which generate the task tree planning for the given goal node(recipe). The knowledge representation of the dishes is called FOON. It contains the different objects and their between them with respective to the motion node The graphical representation of FOON is made by noticing the change in the state of an object with respect to the human manipulators. We will explore how the FOON is created for different recipes by the robots. Task planning contains difficulties in exploring unknown problems, as its knowledge is limited to the FOON. To get the task tree planning for a given recipe, the robot will retrieve the information of different functional units from the knowledge retrieval process called FOON. Thus the generated subgraphs will allow the robot to cook the required dish. Thus the robot can able to cook the given recipe by following the sequence of instructions.

## I. INTRODUCTION

In the previous work, the recipes were prepared with the given set of instructions and by using ingredients in the kitchen. The output of each recipe is the JSON file, which contains the different functional nodes. Each recipe contains multiple functional nodes to prepare the given dish. Each recipe JSON file is considered a subgraph.

An advanced graphical representation is used to visualize the process of making a recipe[1],[2],[3]. The graphical representation is called a Functional Object-Oriented Network (FOON). FOON contains all the activities required for cooking a dish. A FOON is built with different functional units and each functional unit contains the input nodes, output nodes, and motion node.

Each FOON is considered a subgraph. The subgraphs of all the recipes are combined and created as a Universal FOON. This Universal FOON is used to create the different recipes.

Creating this Universal FOON and the subgraphs is the main motivation for developing this algorithm, as this will help the robot to make any dish within its knowledge scope.

## II. FOON CREATION

The smallest unit of FOON is the functional unit which contains the different objects and their states. A functional unit shows one activity in the cooking process.

The set of all functional units for a given dish makes the subgraph. The functional nodes start and end with "// ". For example, look at the functional node in the below fig1.

```
//
O    cutting board
S    contains    {sweet potato}
O    sweet potato
S    peeled
S    in [cutting board]
O    knife
M    cut
O    sweet potato
S    chopped
S    in [cutting board]
//
```
Fig.1 Functional node of sweet potato

The above functional node of sweet potato has different objects and their states.

**Objects:** The objects are represented with 'O'. The above sweet potato functional node contains three objects in the input node cutting board, sweet potato, and knife.

**States:** States are represented with 'S'. The state of the sweet potato object in fig1 is peeled and it is present in a container called cutting board.

From fig1 the Functional node of a sweet potato contains 3 nodes that are input node, motion node, and output node.

**Input node:** The objects that are present above the motion node M are called the input nodes. The input node has 3 objects in fig1.

**Motion node:** The motion node is represented with "M". Each functional unit has only one motion node.

Each Motion node performs certain actions. The action performed by the robot on to the environment is represented by the motion node.

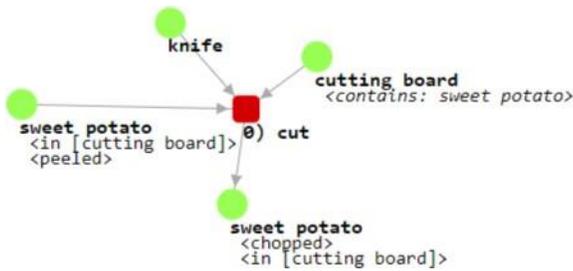

Fig.2 Functional unit of Sweet Potato.

The visual representation of Fig.1 is shown in Fig.2. The visual representation contains nodes and edges.

**Nodes**: The nodes are represented with green circles. The difference between the input and output nodes can be identified by using the edges. For input nodes, the edges go towards the motion node.

For output nodes, the edge goes away from the motion node. Each object node in the FOON is unique[1]. No two objects will have all attributes the same. An object can appear at different locations in the FOON, but it will states will be different.

**Edges:** The link between the nodes is represented by the directed edges. The total number of directed edges in the functional node represents the total number of objects.

A FOON is simply a directed graph. FOON represents the different nodes and edges represent the interaction between those objects[1].

**Task Tree:** The task tree contains different actions that are performed to prepare the dish. The visualization of the task tree contains multiple motion nodes, input nodes, and output nodes.

**Task tree creation:**
The different functional units of a recipe are combined to form a task tree.

By using the FOON, the robot tries to solve the given problem. The robot can generate a task tree that contains the functional units(actions). These actions must be performed to reach the goal node stage. Knowledge retrieval is used to find the task tree in the FOON[2].

A task tree can combine the instructions from different resources to generate the sequence of steps required in recipe preparation. So in gathering instruction from different resources, the algorithm is required to select the optimal functional units based on different criteria.

In this search, we are going to start from the goal node and we gather all the candidate keys in a list. Then we need to select one optimal key among all candidate keys. The optimal key selection process is done according to the algorithm we are developing. Below fig.3 shows the task tree of sweet potato. Fig.3 contains 3 motion nodes and 7 objects.

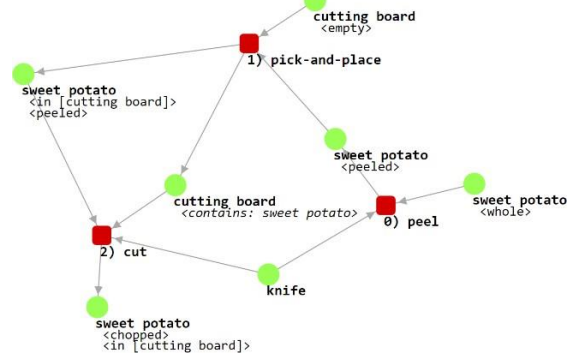

Fig.3 Task tree of Sweet potato

### III. METHODOLOGY

The model is developed by combining different files. The files needed to generate the task tree are Goal_node.json, Kitchen.json, FOON.txt, Utensils.txt, Preprocess.py, Search.py Below is the description of each file.

1) Goal_node.json – This file contains the list of goal nodes, for which the model needs to generate the task tree. This file contains the characteristics like label, states, ingredients, and container.

2) Kitchen.json – This file contains the list of items present in the kitchen with their characteristics.

3) FOON.txt – This is the file, which is containing the knowledge about how to create task planning. The knowledge

of the robot is limited to this FOON.txt file. It contains the recipe data of all students.

4) Utensils.txt – This text file contains the list of utensils available in the kitchen.

5) Preprocess.py – This is a Python file that is used to preprocess the given data in the FOON.txt. This file is used to preprocess the FOON.txt file and also to create the pickle file of FOON.

6) Search.py – This file is used to generate the Task tree. In this file, the algorithm is developed.

This model is developed by using three different algorithms, each has its own condition for selecting the nodes in the graph.

1. **Iterative Deepening Search** – Iterative deepening search is the same as Depth First Search(DFS). But the main difference is that it explores the nodes in the task tree iteratively from the root node to the leaf nodes(Items present in the kitchen.json file) by increasing the depth by 1 for each iteration. Iterative deepening search will explore the node within the depth bound.

2. **Greedy Best-First Search** - In Greedy Best-First Search, the child nodes are not explored randomly instead they are explored by predefined heuristic functions as given below,

    1) **Heuristic 1**:

    Heuristic1(h(n)) is based on the success rate of the motion node. Each functional node contains input, motion, and output. Heuristic1 starts exploration from the root node and it explores the rest of the nodes in the tree based on the success rate of that motion node.

    For example, with Heuristic1 robot tries to choose motion pour rather than pick-and-place, because motion pour has a motion probability of 0.90 whereas motion pick-and-place has a motion probability of 0.80.

    2) **Heuristic 2**:

    Heuristic2(h(n)) explores the nodes, which has the least number of inputs. Firstly, the developed algorithm searches all the functional nodes with the given goal node as the output. Then it explores the functional node that has the least number of inputs.

For example, omelette can be prepared by adding onions, but that is an optional choice. So, the robot chooses the functional node which has less number of inputs such as the input which is without onions.

IV. DISCUSSION

In conclusion, this model has provided different files which contain information about the preparation of recipes with the list of items present in the kitchen. The model has loaded all the given files into search.py and the given algorithm has explored each task tree recipe in different ways according to certain criteria. The FOON.txt file is used by the developed algorithm to generate the task tree planning.

The developed algorithm outputs different task tree plannings for the same recipe. So we need to compare all 3 different outputs for the given same recipe. Tabel 1 perfectly shows the number of output lines the given recipes generate for different algorithms.

Table1 – number of output lines of given recipes

| Goal Nodes | Algorithms | | |
|---|---|---|---|
| | IDFS | H1 | H2 |
| Greek salad | 378 | 490 | 380 |
| macaroni | 99 | 113 | 109 |
| Whipped cream | 187 | 196 | 267 |
| Ice | 19 | 17 | 17 |
| Sweet potato | 30 | 30 | 30 |

The below Table2 shows the number of functional units the given algorithms generate for the given recipes.

| Goal Nodes | Algorithms | | |
|---|---|---|---|
| | IDFS | H1 | H2 |
| Greek salad | 28 | 35 | 28 |
| macaroni | 7 | 7 | 8 |
| Whipped cream | 7 | 10 | 15 |
| Ice | 1 | 1 | 1 |
| Sweet potato | 3 | 3 | 3 |

The above table shows that no particular algorithm is best suited for all recipes. One algorithm works better with one recipe and the other algorithm works better with another recipe.

So, in future research, the algorithm that works fine with all the recipes can be developed based on these results.

**References:**

[1] D. Paulius, Y. Huang, R. Milton, W. D. Buchanan, J. Sam and Y. Sun, "Functional object-oriented network for manipulation learning," 2016 IEEE/RSJ International Conference on Intelligent Robots and Systems (IROS), Daejeon, Korea (South), 2016, pp. 2655-2662, doi: 10.1109/IROS.2016.7759413.

[2] M. S. Sakib, D. Paulius and Y. Sun, "Approximate Task Tree Retrieval in a Knowledge Network for Robotic Cooking," in IEEE Robotics and Automation Letters, vol. 7, no. 4, pp. 11492-11499, Oct. 2022, doi: 10.1109/LRA.2022.3191068.

[3] D. Paulius, A.B Jelodar and Y. sun, "Funcitoal object-Oriented Network: Construction & Expansion," 2018 IEEE International Conferences on Robotics and Automation(ICRA) Brisbane, QLD, Australia, 2018, pp. 5935-5491, doi: 10.1109/ICRA.2018.8460200.